\documentclass{article}

\usepackage[preprint]{neurips_2025}

\usepackage[utf8]{inputenc} 
\usepackage[T1]{fontenc}    
\usepackage{hyperref}       
\usepackage{url}            
\usepackage{booktabs}       
\usepackage{amsfonts}       
\usepackage{nicefrac}       
\usepackage{microtype}      
\usepackage{xcolor}         
\usepackage{amsmath}
\usepackage{amssymb}
\usepackage{algorithm}
\usepackage{algorithmic}
\usepackage{array}
\usepackage{graphicx}
\usepackage{subfig}
\usepackage{multirow}

\title{PrunePEFT: Iterative Hybrid Pruning for Parameter-Efficient Fine-tuning of LLMs}

\author{
  Tongzhou Yu$^{\ast,1}$,\ Zhuhao Zhang$^{\ast,1}$ \\ 
  \textbf{Guanghui Zhu$^{\dagger,1}$, Shen Jiang$^{1}$, Meikang Qiu$^{2}$, Yihua Huang$^{1}$}  \\
  $^{1}$State Key Laboratory for Novel Software Technology, Nanjing University, Nanjing, China \\
  $^{2}$School of Computer and Cyber Sciences, Augusta University, Augusta, Georgia, USA \\
  \texttt{tzyu@smail.nju.edu.cn} \quad \texttt{zhuhaozhang@smail.nju.edu.cn} \\
  \texttt{zgh@nju.edu.cn}$^{\dagger}$ \quad \texttt{jiangshen@smail.nju.edu.cn} \\
  \texttt{qiumeikang@yahoo.com} \quad \texttt{yhuang@nju.edu.cn} \\
}

\begin{document}

\maketitle


\begin{abstract}
Parameter Efficient Fine-Tuning (PEFT) methods have emerged as effective and promising approaches for fine-tuning pre-trained language models.
Compared with Full parameter Fine-Tuning (FFT), PEFT achieved comparable task performance with a substantial reduction of trainable parameters, which largely saved the training and storage costs. 
%
However, using the PEFT method requires considering a vast design space, such as the type of PEFT modules and their insertion layers.
Inadequate configurations can lead to sub-optimal results.
Conventional solutions such as architectural search techniques, while effective, tend to introduce substantial additional overhead.
In this paper, we propose a novel approach, PrunePEFT, which formulates the PEFT strategy search as a pruning problem and introduces a hybrid pruning strategy that capitalizes on the sensitivity of pruning methods to different PEFT modules. 
This method extends traditional pruning techniques by iteratively removing redundant or conflicting PEFT modules, thereby optimizing the fine-tuned configuration. 
By efficiently identifying the most relevant modules, our approach significantly reduces the computational burden typically associated with architectural search processes, making it a more scalable and efficient solution for fine-tuning large pre-trained models.
\end{abstract}

\section{Introduction}

Parameter-efficient fine-tuning (PEFT) has emerged as a prominent paradigm for adapting pre-trained models to downstream tasks while preserving computational efficiency, particularly large language models (LLMs). These methods typically involve the introduction of compact adapter modules~\cite{Houlsby2019adapter}~\cite{he2022unified}, selective parameter freezing or tuning~\cite{zaken2021bitfit}, or leveraging low-rank decomposition techniques to reduce the number of trainable parameters~\cite{hu2021lora}. Notably, PEFT approaches can achieve performance comparable to full fine-tuning, despite training~\cite{Houlsby2019adapter} only a minuscule fraction—often as low as 1\%—of the full parameter set.
\let\thefootnote\relax\footnote{$^{\ast}$These authors contributed equally to this work.}
\let\thefootnote\relax\footnote{$^{\dagger}$Corresponding author}

A myriad of PEFT methods have been proposed~\cite{hu2021lora}~\cite{Houlsby2019adapter}~\cite{zhang2023adalora}, all with the overarching objective of optimizing task performance while minimizing the parameter budget. However, the efficacy of these methods varies significantly across different tasks, as they exhibit distinct sensitivities to task complexity and the allocated parameter budget~\cite{he2021towards}. Previous empirical investigations have also revealed that no single PEFT method universally outperforms others across all tasks~\cite{ding2022largescale}, highlighting the necessity of tailoring the approach to specific task characteristics. 
Moreover, most existing PEFT applications are restricted to a singular method (e.g., Adapter, LoRA), leading to configurations that may not be optimal for a diverse array of tasks. Solving these challenges manually for each individual task is not only labor-intensive but also computationally inefficient.
To address these limitations, researchers have developed versatile PEFT modules capable of adapting to specific tasks, such as AdaLoRA~\cite{zhang2023adalora} and AdapterFusion~\cite{pfeiffer2020adapterfusion}, though the optimal performance and applicability of these modules are often non-intuitive.

In recent advancements, dynamic PEFT configuration strategies have garnered attention. \cite{ruckle2020adapterdrop} tackle this challenge by randomly dropping low-level adapters, resulting in only a marginal reduction in task performance. \cite{zhou2024autopeft} and \cite{hu2022sparse} extend this concept by integrating neural architecture search (NAS) and differentiable architecture search (DARTS)-based methods for fine-tuning strategy discovery, utilizing multi-objective Bayesian optimization to identify superior PEFT configurations. However, these methods typically suffer from high time complexity and significant computational overhead as the model depth and the number of PEFT modules increase.
\citet{lawton2023neural} integrates pruning strategies with NAS, but their approach is limited by a relatively constrained search space and a fixed pruning strategy, while still incurring substantial computational overhead.

To efficiently navigate the vast search space and identify the most effective fine-tuning configurations, we propose an adaptive hybrid pruning-based search strategy called PrunePEFT~\footnote{Code is available at \url{https://github.com/tzwo/PrunePEFT}}. Given the immense parameter space of modern models, the introduction of additional PEFT modules represents a minimal increment in terms of overall parameter count.
Our approach initializes the model by incorporating all potential PEFT modules across each layer and subsequently prunes less impactful modules. 
Drawing on a comprehensive review of pruning strategies~\cite{molchanov2016pruning}, we introduce a hybrid pruning method that merges the advantages of multiple pruning strategies during the warm-up phase, effectively enhancing the quality of the search results. By leveraging this hybrid approach, we can progressively prune the candidate PEFT modules and identify the most efficient fine-tuned configurations under the specified parameter budget.
Our contributions are summarized as follows: 

\textbf{New Problem Formulation:} We formulate the PEFT strategy search as a pruning problem. 
    We first construct a supernet by incorporating potential PEFT modules and then prune the supernet progressively. 
    This formulation can effectively mitigate computational complexity while maintaining robust search performance.

\textbf{Adaptive Hybrid Pruning Strategy:} Based on the observation that different pruning strategies have specific tendencies towards different PEFT modules, we propose an adaptive hybrid pruning strategy to iteratively remove inappropriate PEFT modules for each transformer layer, which can significantly enhance the performance compared to traditional methods relying on a single pruning approach.

\textbf{SOTA Performance and Computational Efficiency.}: Extensive experiments conducted on the multitasking benchmark GLUE \cite{wang2018glue} demonstrate the efficacy of our method. PrunePEFT successfully identifies fine-tuned configurations that match or surpass the performance of full fine-tuning, while only introducing additional overhead of approximately 30\% training time.

\section{Related Work}
The related work can be categorized into three aspects: Parameter Efficient Fine-tuning, Automated PEFT, and pruning strategies. 
Our research shares objectives with PEFT, which aims to adapt pre-trained models to downstream tasks through selective and efficient parameter updates. Furthermore, balancing parameter efficiency with computational cost remains a key challenge in PEFT, which our work explicitly addresses. Automated fine-tuning constructs models through automated frameworks, with related works like \cite{zhou2024autopeft}, \cite{hu2022sparse}, and \cite{jin2019auto} optimizing the design of PEFT configurations using automated methods.
In contrast, our pruning-based fine-tuning strategy employs multiple pruning strategies that have been proven effective in traditional machine learning~\cite{vadera2022methods}. We aim to demonstrate the effectiveness of these methods in the field of PEFT.
\paragraph{Parameter Efficient Fine-tuning.}
Due to massive trainable parameters and large storage requirements, \cite{houlsby2019parameter} proposes parameter-efficient fine-tuning methods, where adapters enable transfer learning by adding additional bottleneck modules. \cite{hu2021lora} uses low-rank decomposition to reduce the number of parameters significantly. Many ingenious variants of adapters and LoRAs have emerged, such as adapter fusion~\cite{pfeiffer2020adapterfusion}, Mad-X~\cite{pfeiffer2020mad}, and AdaLoRA~\cite{zhang2023adalora}.
\cite{mao2021unipelt} mixes PEFT modules via a gating mechanism. \cite{wang2022adamix} creates a mixture of Adapter modules. \cite{liu2024dora} decomposes pre-trained weights into magnitude and direction components. \cite{he2022unified} unifies various adapter and LoRA variants, indicating that the current ingenious designs are combinations of specific components.

\paragraph{Automated PEFT.}

Automated machine learning (AutoML) has received increasing attention recently~\cite{jin2019auto}.
Recent efforts have explored combining PEFT with AutoML. \cite{hu2022sparse} combines PEFT modules by probabilistic gating and differentiable search. \cite{lawton2023neural} employs structured and unstructured pruning to learn PEFT configurations. \cite{zhou2024autopeft} optimizes the model structure with Bayesian algorithm. 

\paragraph{Pruning Strategies.}
Applying structured or unstructured pruning to neural networks is a well-established method. \cite{wen2016learning} introduces group-wise pruning on channels. \cite{molchanov2016pruning} utilizes Taylor's formula as the pruning strategy. \cite{frankle2018lottery} and \cite{malach2020proving} hypothesize and try to verify the correctness of neural network pruning. \cite{morcos2019one} demonstrates the feasibility of the pruning strategy on NLP tasks and further discovers that the winning tickets from large-scale datasets are applicable to small datasets. \cite{Cheong2019transformersZ} prunes the Transformer model to reduce the number of parameters. \cite{sajjad2020poor} prunes 40\% of the model's weights while having a minimal influence on performance.
\cite{shen2022prune} compares post-pruning and pre-pruning and introduces early training epochs which are suitable for pruning. 

We adopt pruning strategies such as weight-based pruning ~\cite{han2015learning}, zero-count-based pruning ~\cite{hu2016network}, and Taylor-expansion-based pruning ~\cite{molchanov2016pruning}.

\section{Motivation}
\paragraph{PEFT Module Search as Network Pruning.}  
Traditional approaches to optimizing sub-networks or hyperparameter configurations often involve search methodologies such as neural architecture search. These methods typically require multiple rounds of training and evaluation, resulting in substantial computational overhead and significant time consumption. While strategies like limiting the search time or employing low-fidelity optimization can enhance efficiency, the overall computational cost remains high. In contrast, pruning strategies iteratively shrink a super-network to identify optimal sub-networks. This approach inherently imposes an upper bound on computational resource consumption, which is substantially lower than direct search methods. Particularly in simplified search spaces, such as those for SA(serial adapter, e.g., Adapter) and PA(parallel adapter, e.g., LoRA) , if model contains $n$ PEFT modules, pruning strategies can achieve time complexity as low as $O(n)$.  

\begin{table}[t]
  \caption{Pruning strategies focus on information from multiple aspects of model computation.}
  \label{table-1}
  \centering
  \begin{tabular}{c|c|c|c}
    \toprule
    \textbf{Focus} &\textbf{Method} & \textbf{Standard} & \textbf{Formulas} \\
    \midrule
    \multirow{3}{*}{FC} 
    &Weight& Mean squared magnitude of weights & $\Theta(m_i) = \frac{1}{|\boldsymbol{w}|} \sum_{i} w_i^2$ \\
    &Zeros& Proportion of non-zero weights & $\Theta(m_i) = \frac{1}{|\boldsymbol{w}|} \sum_{i} \mathbb{I}(w_i \neq 0)$ \\
     &Threshold& Proportion of weights below a threshold $t$ &$\Theta(m_i) = \frac{1}{|\boldsymbol{w}|} \sum_{i} \mathbb{I}(|w_i| < t)$ \\
    \midrule
    IoF
    &Activation& Average activation magnitude & $\Theta(m_i) = \frac{1}{|\boldsymbol{w}|} \sum_{i} |f(w_i)|$ \\
    \midrule
    \multirow{2}{*}{IoG}
    &Sensitivity& Taylor expansion-based importance  & $\Theta(m_i) = \frac{1}{|\boldsymbol{w}|} \sum_{i} \left| w_i \frac{\partial \mathcal{L}}{\partial w_i} \right|$ \\
    &Gradient& Mean absolute gradient magnitude & $\Theta(m_i) = \frac{1}{|\boldsymbol{w}|} \sum_{i} \left| \frac{\partial \mathcal{L}}{\partial w_i} \right|$ \\
    \bottomrule
  \end{tabular}
\end{table}

Prior research highlights that hidden layers in neural networks encode distinct types of information~\cite{vulic2020probing}, with lower layers capturing general surface-level information, while deeper layers are more task-specific ~\cite{tenney2019bert}. This allows for partitioning the model into distinct segments, enabling the application of varied PEFT methods across these partitions ~\cite{chen2023designspaces}.  

Pruning techniques leverage diverse criteria to identify and retain critical modules, including activation values, gradient information, and parameter-specific characteristics , etc~\cite{molchanov2016pruning}. These criteria correspond to different stages of model computation: feature characteristics (FC), information about the forward propagation stage (IoF), and information about gradients from backpropagation (IoG). Each of these strategies provides unique insights into model behavior, and their formulas are detailed in Table \ref{table-1}.  
\begin{figure*}[t]
\centering
\subfloat[]{
		\includegraphics[scale=0.35]{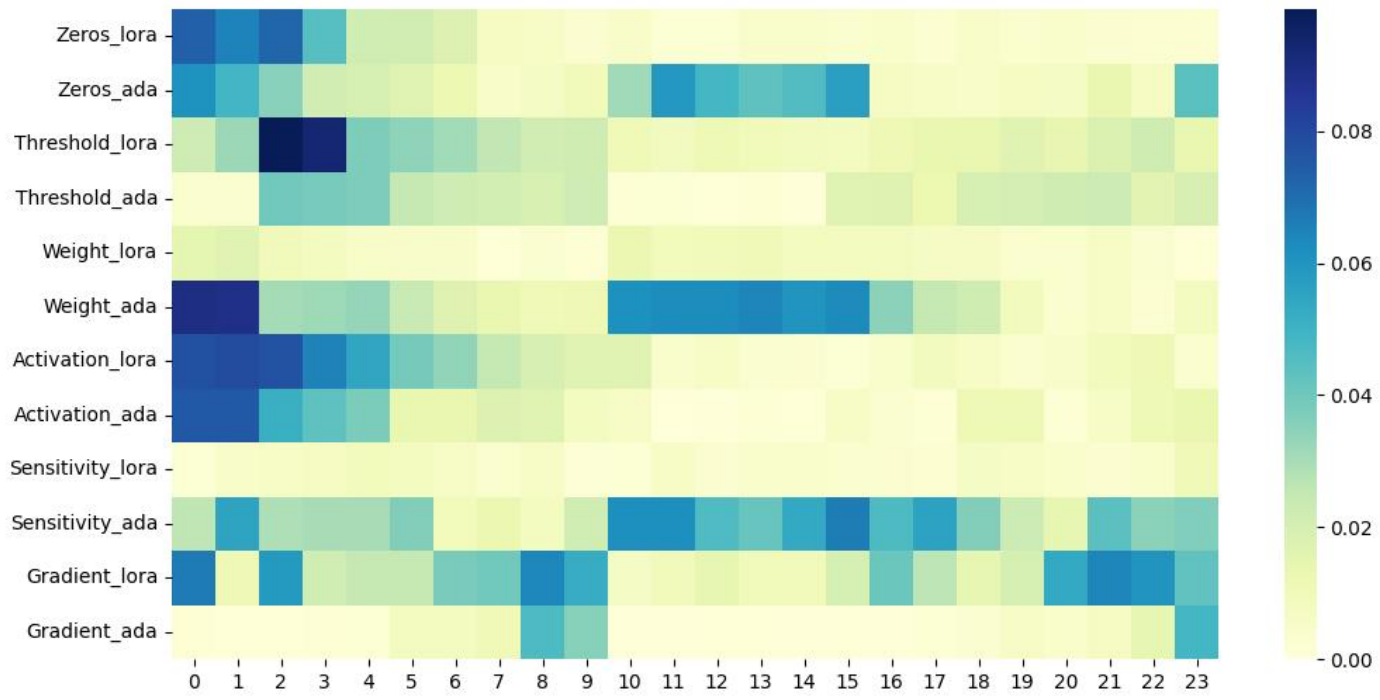}}
\subfloat[]{
		\includegraphics[scale=0.455]{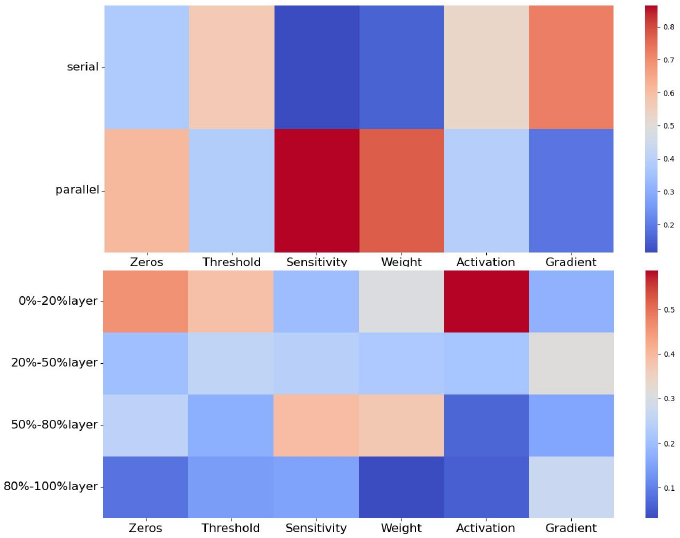}}
\caption{The visualizations illustrate the sensitivities of distinct pruning strategies to the selection of PEFT modules. (a) depicts the pruning ratios of various strategies applied to different PEFT modules, with the horizontal axis representing layer indices. (b) summarizes the results: 1) The pruning tendencies of different strategies for various PEFT methods, and 2) The pruning tendencies of different strategies for PEFT modules in different model layers.}
\label{figure-1}
\end{figure*}

Our experimental findings corroborate previous research~\cite{he2021blending} that different pruning strategies yield inconsistent rankings of layer importance. Figure \ref{figure-1} highlights the pruning tendencies of various strategies, leading to the following observations:

\textbf{Observation 1: Pruning strategies demonstrate PEFT method-specific preferences.}  
Distinct pruning strategies show varying tendencies towards specific PEFT modules. For instance, the Taylor expansion-based strategy tends to preferentially prune the serial modules within the PEFT framework, emphasizing its sensitivity to PEFT configurations rather than solely to depth.

\textbf{Observation 2: Pruning strategies exhibit varying sensitivities to model depth.}  
Pruning strategies display distinct preferences for layers at different depths within the model. For example, in tasks derived from the GLUE dataset, the activation-based pruning strategy primarily targets layers located in the top 20\% of the model, suggesting a stronger sensitivity to upper-layer representations.

Each pruning strategy selects and removes less significant components of the model based on specific, valid information sources to enhance performance. To further improve the quality of the searched fine-tuned structures, we propose a unified hybrid pruning strategy that organically integrates insights from these diverse information sources. 

\begin{figure*}[ht]
\centering
\includegraphics[width=1\textwidth]{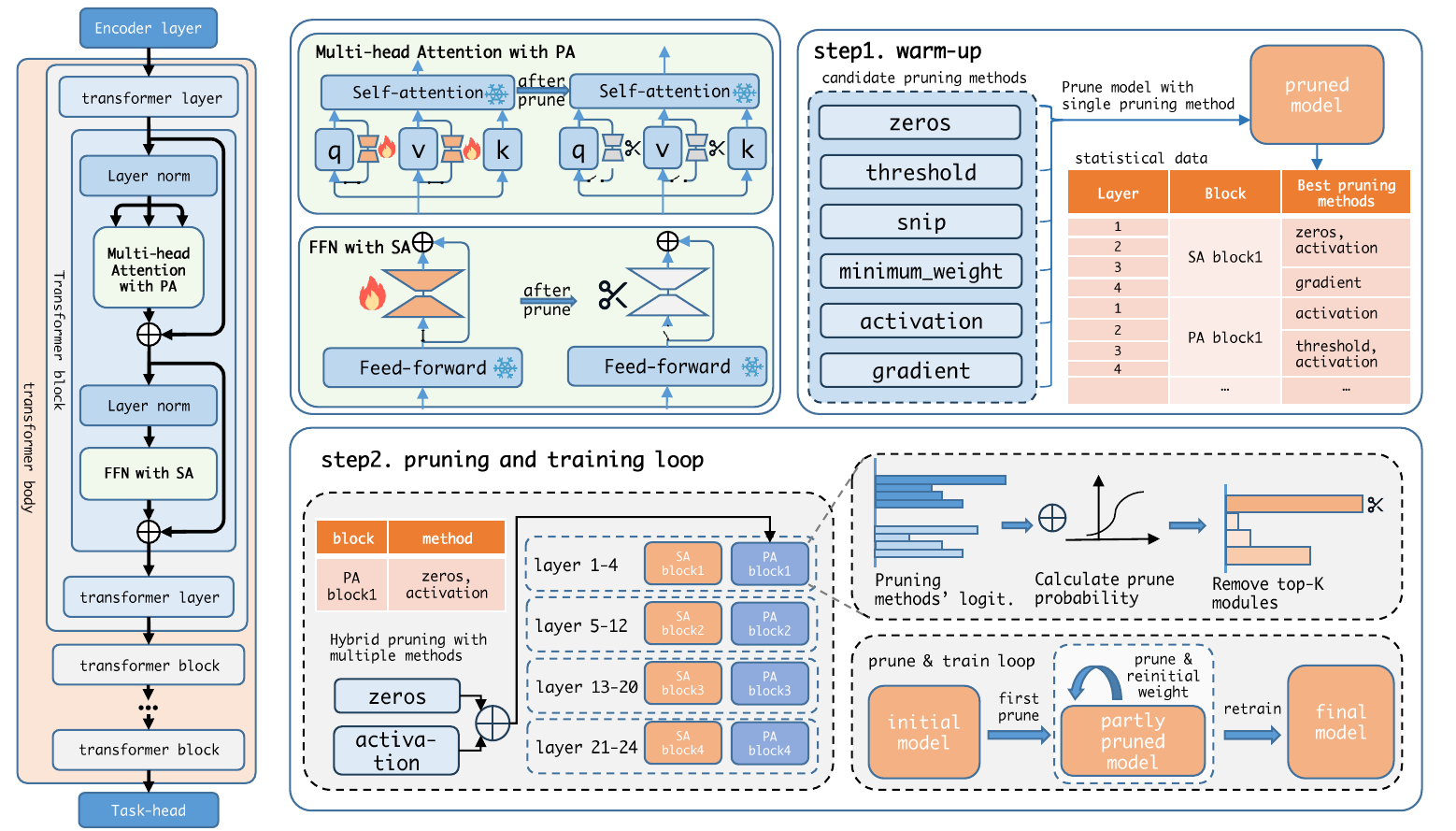}
\caption{The overall framework of the proposed PrunePEFT (i.e., iterative hybrid pruning), which consists two phases: warm-up phase and iterative pruning phase. 
}
  \label{figure-2}
\end{figure*}

\section{The Proposed Method: PrunePEFT}
The goal of this work is to identify the optimal fine-tuning configuration $M$ under the constraint of a predefined and finite budget $B$ of trainable parameters. 
In this paper, we propose a novel iterative hybrid pruning approach called PrunePEFT to eliminate less important PEFT modules. 
As shown in Figure~\ref{figure-2}, the overall framework consists of two primary phases: (i) warm-up phase that determines the set of hybrid pruning strategies for each mode partition, and (ii) iterative pruning phase that achieves iterative hybrid pruning tailored to downstream tasks.
After multiple rounds of pruning, we retrain the 
searched fine-tuned configuration.
The proposed method enables both the architectural search and the control of sparsity in PEFT modules.

\subsection{Iterative Hybrid Pruning}
\paragraph{Hybrid Pruning Strategy Selection with Warm-up.}  
To identify the most critical PEFT modules, we select pruning strategies based on various sources of information, thereby employing a diverse set of pruning strategies, as outlined in Table~\ref{table-1}. 

The model is partitioned into four spindle-shaped blocks~\cite{chen2023designspaces}, with a warm-up phase introduced prior to training. During the warm-up phase, the original dataset $D$ is sampled to generate a compact dataset $d$, which maintains the original label distribution. 
The model then undergoes multiple rounds of heuristic training on $d$. Throughout these training iterations, we track the pruning frequency of each pruning strategy for various PEFT modules, such as parallel components located in the first block of the model. This pruning frequency serves as the pruning tendencies for each strategy. Based on these observed tendencies, we assign the pruning strategy $S_i$ that most effectively matches the characteristics of partition $P_i$.  
The pruning strategy $S_i$ of each partition forms the hybrid pruning strategy $S_H$.

For partition $P_i$, we employ Bayesian model averaging to jointly determine the pruned layers. Specifically, the probability $p(m_i)$ of pruning module $m_i$ is computed as follows:
\begin{equation}
\begin{aligned}
    p(m_i) &= \frac{e^{k_i \cdot \Theta(m_i, S_i)}}{\sum_{j=1}^{n} e^{k_j \cdot \Theta(m_j, S_j)}}, \\
    \text{where} \quad k_i &= \text{Sigmoid}(rank(m_i, S_i)).
\end{aligned}
\end{equation}
Here, $\Theta(m_i, S_i)$ represents the evaluation metric of PEFT module $m_i$ under pruning strategy $S_i$, and $k_i$ is the weight determined by the rank of the module’s likelihood of being pruned by strategy $S_i$. The hybrid pruning strategy $S_H$ for the entire dataset $D$ is derived by combining the pruning strategies from all partitions.
\paragraph{Iterative Pruning.}  
To obtain the final fine-tuned configuration $M$, we apply the iterative pruning approach based on the hybrid pruning strategy $S_H$ identified during the warm-up phase. In each round of pruning, the model is trained in several steps using either the full dataset $D$ or the trimmed dataset $d$, with both forward and backward propagation data (e.g., gradients) recorded. During each pruning round, the hybrid pruning strategy $S_H$ computes the probability of pruning each module based on intermediate information. The modules with the highest $k$ pruning probabilities are removed from the model:
\begin{align}
    \text{mask} = & \ \text{Onehot}(TopK(P,k))\\
    \quad M_i &= \text{mask}(M_{i-1})
\end{align}
where $P$ denotes the importance ranking based on the hybrid pruning strategy $S_H$, and $M_i$ represents the model after the $i$-th round of pruning. Each subsequent round of pruning operates on the model $M_i$ obtained from the previous iteration. If the current round is not the last, the pruned PEFT modules are reinitialized as follows~\cite{lawton2023neural}:
\begin{align}  
    W_{down} &\sim \mathcal{N}(0, 1/\sqrt{d_{down}}), \\ W_{up} &\sim \mathcal{N}(0, 1/\sqrt{d_{up}}),
\end{align}
where $W_{down}$ and $W_{up}$ correspond to the down-projection and up-projection matrices, respectively, and $d_{down}$ and $d_{up}$ denote their corresponding dimensions.  

After completing the predetermined number of pruning rounds, the total number of trainable parameters reaches the target budget $B$. At this point, the remaining PEFT modules are considered the final fine-tuned configuration $M^*$ for the task at hand. The output of layer $h_i$ is then computed as:
\begin{align}
    H_{\text{out}} =\text{mask}^{SA}_{h_i} \cdot \left( \sigma(H_{\text{in}} \cdot W^{SA}_{\text{down}}) \cdot W^{SA}_{\text{up}} \right)  + \text{mask}^{PA}_{h_i} \cdot \left( \sigma(H_{\text{in}} \cdot W^{PA}_{\text{down}} \cdot W^{PA}_{\text{up}}) \right)
\end{align}
where $\text{mask}^{SA}_{h_i}$ and $\text{mask}^{PA}_{h_i}$ belonging to the $\text{mask}_{h_i}$ are boolean indicators that control whether the corresponding PEFT modules are activated at layer $h_i$. $\sigma(\cdot)$ is the nonlinear activation function such as ReLU or GELU.
The overall workflow of iterative hybrid pruning is shown in Algorithm~\ref{alg:pruning}.

\begin{algorithm}[t]
\caption{Iterative Hybrid Pruning}  
\label{alg:pruning}
\begin{algorithmic}[1]
\STATE \textbf{Input:} Dataset $D$, initial model $M$ with all PEFT modules, number of pruning rounds $r$
\STATE \textbf{Output:} Final pruned model $M^*$
\STATE \textbf{Warm-up Phase:}
    \STATE Train the model on the trimmed dataset $d \subseteq D$. For each partition $P_i$ of the model, calculate the pruning strategy $S_i$ using heuristic methods. Based on the observed tendencies, assign the appropriate pruning strategy to each partition to form the hybrid pruning strategy $S_H$.
\STATE \textbf{Iterative Pruning Phase:}
\FOR{$i = 1$ \TO \texttt{r}}
    \STATE Compute the pruning probability vector $P$ for each module using the hybrid pruning strategy $\Theta(M_{i-1}, S_H)$.
    \STATE Identify and remove the top $k$ layers based on the pruning probability $P$: $\text{mask} = \text{Onehot}(\text{TopK}(P, k))$. Update the model: $M_i = \text{mask}(M_{i-1})$.
    \STATE Reinitialize the pruned layers: $W_{down} \sim \mathcal{N}(0, 1/\sqrt{d_{down}})$, $W_{up} \sim \mathcal{N}(0, 1/\sqrt{d_{up}})$.
\ENDFOR
\STATE \textbf{Return:} The pruned model $M^*$ after $r$ rounds, representing the final configuration.
\end{algorithmic}
\label{alg}
\end{algorithm}
\subsection{Complexity Analysis}
For our search space, the total number of possible configurations is given by $(N_{PA} * N_{SA})^{|H|}$, where $N_{PA}$ and $N_{SA}$ represent the number of choices for PA and SA methods at each layer, respectively, and $|H|$ denotes the number of layers in the model. During a forward search, evaluating each configuration requires $r$ epochs, with each epoch taking time $t$, leading to substantial computational overhead. PrunePEFT mitigates this challenge by using the pruning approach. In each pruning round, after one training epoch to gather necessary information, $k$ modules of lower importance are pruned. Given that, in the worst case, all of PEFT modules in model may be eliminated, the total time $T_{search}$ overhead for evaluating a configuration is bounded by:
\begin{align}  
    T_{search} = (N_{PA} * N_{SA}) * ({|H|}/{k}) * t.
\end{align}
 This approach significantly reduces the search time overhead, making it particularly advantageous for search spaces with a larger number of layers and fine-tuning modules.

\section{Experiments}
\subsection{Experimental Settings}
\subsubsection{Datasets and Pre-trained Models (PTMs)}
In line with prior research on parameter efficient fine-tuning, we apply our adaptive hybrid pruning strategy to the multi-task benchmark GLUE \cite{wang2018glue} for the search of fine-tuning configurations.
The GLUE benchmark includes several text classification tasks, namely: Linguistic Acceptability (CoLA), Similarity and Paraphrase (MRPC, QQP), Sentiment Analysis (SST-2), and Natural Language Reasoning (RTE, QNLI, MNLI). 
All datasets are from the HuggingFace Datasets repository \cite{lhoest2021datasets}. Consistent with previous PEFT studies, we use the RoBERTa-large model (326M parameters) as the backbone pre-trained model (PTM).
To demonstrate the scalability of PrunePEFT to larger models, we also evaluate PrunePEFT on Llama3.1-8B.

\subsubsection{Baselines}

We compare the configurations identified by the adaptive hybrid pruning strategy against several common baselines, which include full fine-tuning (FFT) as well as the individual PEFT method present within the search space.

\textbf{Full Fine-Tuning (FFT).} The traditional approach, where all parameters of the pre-trained model are fine-tuned during training.

\textbf{Low-Rank Adaptation (LoRA).} The LoRA approach is applied to the query and value layers of the Self-Attention mechanism as PA modules, following the configuration described in \cite{lawton2023neural}.

\textbf{Adapter.} We adopt the adapter-based method proposed by \citet{Houlsby2019adapter} as another representative PEFT baseline.

We also reproduce the recently introduced DoRA approach \cite{liu2024dora} for comparison.

To further contextualize the performance of our method, we compare it with several existing automated fine-tuning strategies, including S-MaM~\cite{lawton2023neural}, S3Delta~\cite{hu2022sparse}, and AutoPEFT~\cite{zhou2024autopeft}. For fairness, we ensure that both the search space and parameter budget of S3Delta and AutoPEFT are aligned with those used in PrunePEFT.

\begin{table*}
  \caption{\textbf{Results on GLUE Benchmark with RoBERTa-large.} We present the average fine-tuning parameters for the configurations obtained through PrunePEFT across each GLUE task. The reported metrics include Matthew's correlation for CoLA, and accuracy for all other tasks. The parameter percentage denotes the ratio of additional parameters introduced by fine-tuning relative to the number of parameters in the pre-trained model. We replicate some of baseline configurations, with the best performance highlighted in \textbf{bold} and the sub-optimal result highlighted in \underline{underline}.$*$ denotes the results that are directly copied from Hugging face and S-MaM~\cite{lawton2023neural}.
}
  \label{table-2}
  \centering
  \resizebox{1\linewidth}{!}{
  \begin{tabular}{l|l|l|l|l|l|l|l|l|l}
    \toprule
    \textbf{Method} & \textbf{\%Param.} & \textbf{MNLI} & \textbf{QNLI} & \textbf{QQP} & \textbf{RTE} & \textbf{SST-2} & \textbf{MRPC} & \textbf{CoLA} & \textbf{Avg}\\
    \midrule
    FFT* & 100\% & 90.2 & 94.7 & 92.2 & 86.6 & 96.4 & 90.9 & 68 & 88.4\\
    Adapter(r=128) & 2\% & 90.0 & 94.1 & 89.2 & 83.8 & 96.1 & 85.6 & 65.0 & 86.3 \\ 
    LoRA(r=32) & 1\% & 89.9 & 93.8 & 89.8 & 84.0 & 96.0 & 86.3 & 65.3 & 86.4 \\ 
    DoRA(r=32) & 1\% & 90.1 & \underline{94.7} & 90.1 & 83.8 & \textbf{96.9} & 86.0 & 63.8 & 86.5\\ 
    \midrule
    S3Delta & 1\% & {90.1} & \underline{94.7} & \underline{90.4} & {84.5} & 95.8 & \underline{90.2} & {65.8} & {87.4} \\  
    S-MaM* & 1\% & \textbf{90.6} & 94.5 & \textbf{90.6} & \underline{85.2} & 95.9 & \textbf{90.4} & {66.3} & \underline{87.6} \\  
    AutoPEFT & 1\% & {90.3} & \underline{94.7} & \textbf{90.6} & {85.0} & 96.3 & {89.7} & \textbf{66.9} & \underline{87.6} \\  
    \midrule
    PrunePEFT & 1\% & \underline{90.5} & \textbf{{94.9}} &\textbf{90.6} & \textbf{85.5} & \underline{{96.6}} & \textbf{90.4} & \underline{66.6} & \textbf{87.9}\\ 
    PrunePEFT(low fidelity) & 1\% & {90.1} & \textbf{{94.9}} &{90.0} & {85.0} & {96.3} & {88.2} & {66.0} & 87.0\\ 
    \bottomrule
  \end{tabular}
  }
\end{table*}

\begin{table*}
  \caption{\textbf{Results on GLUE Benchmark with Llama3-8B.}  We present the baseline performance results of LoRA and DoRA, evaluated under the same parameter budget. Additionally, we report the performance of the fine-tuned configurations identified by PrunePEFT.
}
  \label{table-3}
  \centering
  \resizebox{1\linewidth}{!}{
  \begin{tabular}{l|l|l|l|l|l|l|l|l|l}
    \toprule
    \textbf{Method} & \textbf{\%Param.} & \textbf{MNLI} & \textbf{QNLI} & \textbf{QQP} & \textbf{RTE} & \textbf{SST-2} & \textbf{MRPC} & \textbf{CoLA} & \textbf{Avg}\\
    \midrule
    LoRA(r=32) & 0.40\% & 88.0 & \underline{95.0} & \underline{90.6} & \textbf{86.6} & 97.0 & \underline{84.2} & 61.9  & 86.2\\ 
    DoRA(r=32) & 0.40\% & \underline{89.8} & 94.7 & 90.1 & \textbf{86.6} & \underline{97.2} & 83.9 & \textbf{66.9} & \underline{87.0} \\
    \midrule
    PrunePEFT & 0.34\% & \textbf{90.2} & \textbf{95.2} &\textbf{90.8} & \underline{84.8} & \textbf{97.3} & \textbf{84.8} & \underline{66.4} & \textbf{87.3}\\ 
    \bottomrule
  \end{tabular}
  }
\end{table*}

\subsubsection{Implementation Details}

For consistency with previous work on the GLUE benchmark, we report the baseline results across all GLUE datasets in Table \ref{table-2}. All experiments are conducted with 20 epochs for training, except where stated otherwise. We employ the codebase from \cite{pfeiffer2020adapterhub}, utilizing the RoBERTa-large model \cite{liu2019roberta} and  Llama3-8B as the backbone PTMs. The batch sizes used in the experiments are 32 and 16, with learning rates set in the range of $(10^{-5}, 10^{-4})$. Multiple configurations are tested to optimize performance on the various tasks within the GLUE benchmark.
\subsection{Results on GLUE}
Table~\ref{table-2} summarizes the performance of our method on various tasks from the GLUE benchmark using RoBERTa-large. Compared to individual PEFT baselines, PrunePEFT consistently achieves superior results across nearly all tasks, and even surpasses full fine-tuning in certain cases. When compared to other automated fine-tuning strategies under an equivalent parameter budget, PrunePEFT also attains the highest average score. Furthermore, we introduce a low-fidelity variant of PrunePEFT, which performs pruning using only a small subset (approximately 1\%) of the original dataset. This significantly reduces the overhead during the search phase, while still retaining 98.4\% of the performance of full fine-tuning.

Table~\ref{table-3} presents the performance of PrunePEFT on the Llama3-8B model. Our findings indicate that PrunePEFT is capable of accurately identifying redundant components within large models. The fine-tuned structure obtained through the search process achieves superior performance with a reduced parameter count.
\subsection{Efficiency of the Search Process}

Traditional architectural search methods typically incur significant time overhead when searching for optimal configurations. To assess the efficiency of our approach, we analyze the ratio of search epochs to retraining epochs during the fine-tuned architecture search on RoBERTa-large. As shown in Table \ref{table-5}, PrunePEFT introduces an additional overhead of less than 30\% of the retraining time for fine-tuned architecture search including the time of the warm-up phase, while PrunePEFT(low fidelity) incurs only 1\% of the retraining time. The search overhead is minimal in comparison to the total training time required for the task, demonstrating the high efficiency of our method.This advantage becomes particularly pronounced when fine-tuning models with larger parameter counts, such as Llama3-8B and beyond.
\begin{table*}[t]
\centering
\renewcommand{\arraystretch}{1.2}
\begin{minipage}[t]{0.47\textwidth}
  \caption{The average ratio of search time to retraining time for the fine-tuned architecture search across the GLUE benchmark tasks.}
  \label{table-5}
  \centering
  \resizebox{1\textwidth}{!}{
  \begin{tabular}{l|c|c|c}
    \toprule
    \multirow{2}{*}{\textbf{Method}} & \multicolumn{3}{c}{\textbf{Average epochs}}\\
    \cline{2-4}
    & \textbf{Search} & \textbf{Retrain} & \textbf{Ratio} \\
    \midrule
    AutoPEFT & 185 & 20 & 9.25 \\
    S3Delta & 130 & 20 & 6.50 \\
    S-MaM & 20 & 20 & 1.00 \\
    PrunePEFT & 6.0 & 20 & \textbf{0.30} \\
    PrunePEFT (low fidelity) & 0.1 & 20 & \textbf{0.01} \\
    \bottomrule
  \end{tabular}}
\end{minipage}
\hfill
\begin{minipage}[t]{0.505\textwidth}
  \caption{Ablation study with different pruning strategies with the RoBERTa-large model under the GLUE benchmark.}
  \label{table-6}
  \centering
  \resizebox{1\textwidth}{!}{
  \begin{tabular}{l|c|c|c}
    \toprule
    \textbf{Method} & \textbf{QNLI} & \textbf{SST-2} & \textbf{MNLI} \\
    \midrule
    Without Prune & 94.1 & 95.8 & 90.0 \\
    Random Prune & 93.8 & 95.6 & 89.5 \\
    Prune with Single Strategy & 94.7 & 96.4 & 90.1 \\
    Without Block Partition & 94.7 & 96.3 & 90.2 \\
    \midrule
    PrunePEFT & \textbf{94.9} & \textbf{96.6} & \textbf{90.5} \\
    \bottomrule
  \end{tabular}}
\end{minipage}
\end{table*}

\subsection{Ablation Study}

We conduct a series of ablation studies to evaluate the effectiveness of the hybrid pruning strategy for structural optimization. The experiments include: a no-pruning baseline, random pruning, single pruning strategy (selected from Table~\ref{table-1}), and hybrid pruning without block partition. Since the optimal single pruning strategy varies across tasks, we report the best-performing result for each task. All experiments are conducted under identical training settings, datasets, and task configurations. As shown in Table~\ref{table-6}, compared to the no-pruning baseline, PrunePEFT improves fine-tuning performance by eliminating redundancy and reducing module interference. Relative to other pruning strategies, the hybrid approach yields more effective model configurations at the end of the search process by extracting richer task-relevant information, resulting in superior performance and greater stability. These findings highlight the effectiveness of the hybrid pruning strategy in optimizing model configurations during the fine-tuning stage.

\begin{table}
  \caption{Transferability evaluation of PrunePEFT with different configurations on RoBERTa-large. }
  \label{table-7}
  \centering
{
  \begin{tabular}{l|l|l|l|l}
    \toprule
\textbf{Source Task} & \textbf{Transfer Type} & \textbf{QNLI} & \textbf{SST-2} &\textbf{MNLI} \\
    \midrule
No Transfer & NaN &{94.9} & {96.6} &{90.5} \\
RTE & Pruning Strategy Transfer&{94.9} & 96.4 & 90.3\\
RTE & Fine-tuned Architecture Transfer&94.7 & 96.4  &90.1\\
CoLA & Pruning Strategy Transfer&{94.9} & 96.4 & 90.3\\
CoLA & Fine-tuned Architecture Transfer&94.8 & 96.3  &90.3\\
    \bottomrule
  \end{tabular}
  }
\end{table}

\subsection{Transferability}
We assess the transferability of the PrunePEFT method by using task RTE and CoLA as the source dataset.
We evaluate the transferability of both the hybrid pruning strategy determined during the warm-up phase, and the fine-tuned architecture derived from the search in the source dataset.
The strategies and configurations are then applied to other datasets for further training. As presented in Table \ref{table-7}, the strategy and architecture from source dataset exhibits strong transferability, whose performance closely matching the performance achieved on the original task. This confirms the reusability of both the hybrid pruning strategy and the fine-tuned architecture, demonstrating the method's adaptability across different datasets.
\begin{figure*}[t]
\centering
\subfloat[]{
		\includegraphics[scale=0.205]{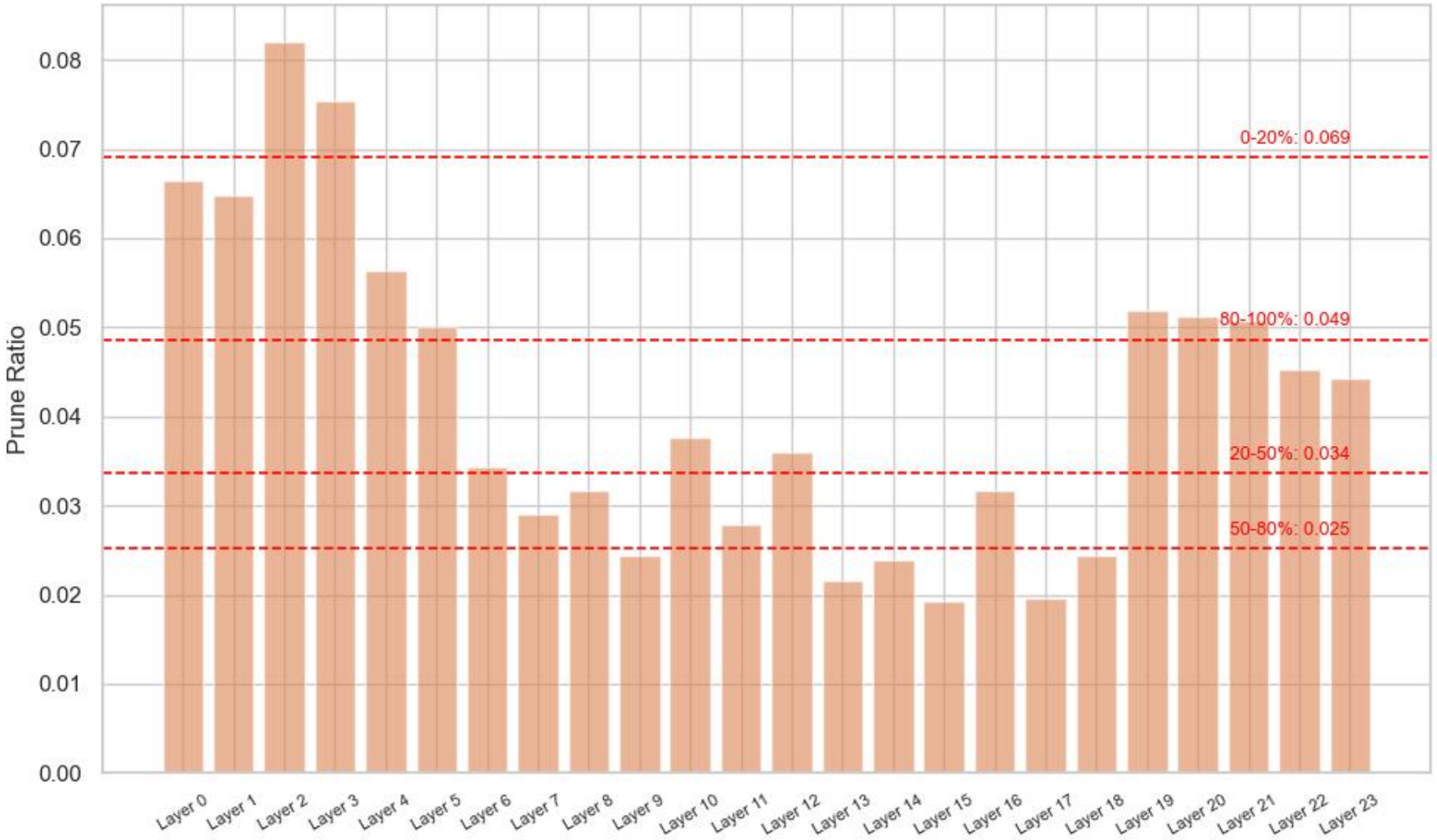}}
\subfloat[]{
		\includegraphics[scale=0.248]{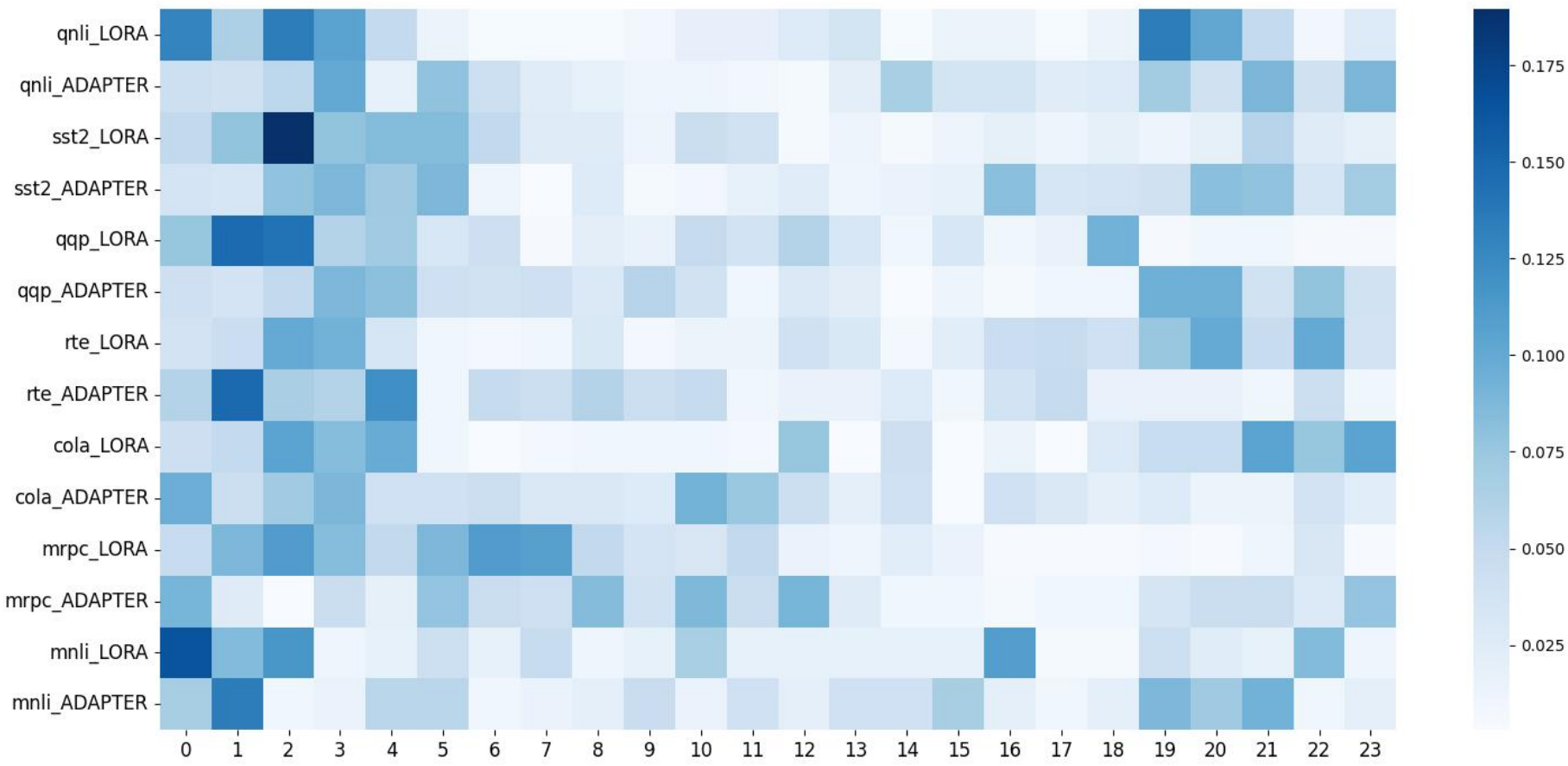}}
\caption{Visualization of the fine-tuned configurations obtained through PrunePEFT. (a) presents the frequency that PEFT module is pruned in diffrent layers and the average frequency in four blocks. (b) presents the frequency with which PEFT module is pruned across different downstream tasks.}
\label{figure-3}
\end{figure*}
\subsection{Explanations of the Searched Structures}
We conduct an analysis of the fine-tuned configurations identified by PrunePEFT and deliberate on the most effective locations for placing PEFT modules within the network. Figure \ref{figure-3} (a) visualizes the fine-tuned configurations obtained through PrunePEFT, providing insights into the placement of PEFT modules. 
Notably, we observe that the PEFT modules deemed unimportant predominantly appear in the initial layers of the model, with a smaller proportion found in the deeper layers and few scattered in intermediate positions. This suggests that fine-tuning layers situated in the middle of the model may yield more effective results.

Regarding the selection of PEFT modules, Figure \ref{figure-3} (b) reveals a distinct, staggered distribution in the frequency with which different PEFT methods are pruned. Specifically, layers where LoRA modules are applied tend to be unsuitable for incorporating Adapter modules, and vice versa. This finding underscores the fact that a single fine-tuning method is not universally applicable across all layers of the model. Consequently, it highlights the importance of performing an architectural search to identify the most suitable configuration for each layer, ensuring optimal model performance.

\section{Conclusion and Future Work}
In this paper, we proposed a novel method for fine-tuning configuration search based on adaptive hybrid pruning, termed PrunePEFT. Experimental results demonstrate that PrunePEFT effectively removes redundant structures, enabling the identification of the optimal PEFT configuration while incurring significantly lower time overhead compared to traditional configuration search methods. There remain several avenues for future work: (1) Exploring more efficient search spaces to enhance the potential of configuration search, and (2) Extending the current pruning strategy to incorporate unstructured pruning techniques, and investigating its impact on fine-tuned configuration search.

\bibliographystyle{unsrtnat}
\bibliography{ref}

\end{document}